\documentclass[conference]{IEEEtran}
\IEEEoverridecommandlockouts
\usepackage{cite}
\usepackage{amsmath,amssymb,amsfonts}
\usepackage{algorithmic}
\usepackage{graphicx}
\usepackage{textcomp}
\usepackage{xcolor}
\usepackage{gensymb}
\usepackage{float}
\usepackage{hyperref}
\usepackage{booktabs}
\usepackage[frozencache,cachedir=.]{minted}
\usepackage[inkscapelatex=false]{svg}
\usepackage{adjustbox}
\usepackage{pdfpages}
\usepackage[font=small,labelfont=bf]{caption}
\usepackage[font=small,labelfont=bf]{subcaption}

\def\BibTeX{{\rm B\kern-.05em{\sc i\kern-.025em b}\kern-.08em
    T\kern-.1667em\lower.7ex\hbox{E}\kern-.125emX}}
\begin{document}

\newcommand{\qq}[1]{TERA}

\title{\qq{}: A Simulation Environment for Terrain Excavation Robot Autonomy\\
}

\author{
Christo Aluckal\textsuperscript{1}, Roopesh Vinodh Kumar Lal\textsuperscript{1}, Sean Courtney\textsuperscript{1}, Yash Turkar\textsuperscript{1}, Yashom Dighe\textsuperscript{1}, Young-Jin Kim\textsuperscript{1}, \\
Jake Gemerek\textsuperscript{2}, Karthik Dantu\textsuperscript{1}
\thanks{\textsuperscript{1} University at Buffalo} 
 \thanks{\textsuperscript{2} MOOG Inc.}
}


\maketitle

\newcommand{\christo}[1]{}
\newcommand{\kar}[1]{}
\newcommand{\yash}[1]{}
\newcommand{\yom}[1]{}
\newcommand{\jake}[1]{}
\newcommand{\sean}[1]{}
\newcommand{\young}[1]{}

\begin{abstract}
Developing excavation autonomy is challenging given the environments where excavators operate, the complexity of physical interaction and the degrees of freedom of operation of the excavator itself. Simulation is a useful tool to build parts of the autonomy without the complexity of experimentation. Traditional excavator simulators are geared towards high fidelity interactions between the joints or between the terrain but do not incorporate other challenges such as perception required for end-end autonomy. A complete simulator should be capable of supporting real-time operation while providing high fidelity simulation of the excavator(s), the environment, and their interaction. 
In this paper we present \qq{} (Terrain Excavation Robot Autonomy), a simulator geared towards autonomous excavator applications based on Unity3D/AGX that provides the extensibility and scalability required to study full autonomy. It provides the ability to configure the excavator and the environment per the user requirements. 
We also demonstrate realistic dynamics by incorporating a time-varying model that introduces variations in the system's responses. 
The simulator is then evaluated with different scenarios such as track deformation, velocities on different terrains, similarity of the system with the real excavator and the overall path error to show the capabilities of the simulation. \qq{} source will be available at \href{https://droneslab.github.io/tera}{https://droneslab.github.io/tera} on publication. 
\end{abstract}


\section{Introduction}



Developing autonomy algorithms for outdoor tasks such as excavation is extremely challenging. Perception in offroad environments where excavation tasks happen is arguably more challenging
than in more structured environments such as roads. Further, task-specific perception involves
understanding soil characteristics, detailed terrain models, and full 360\degree\  awareness for safety. Motion planning for navigation involves reasoning about obstacles as well as terrain elevations as is the case in most offroad scenarios. Task-based planning requires detailed modeling of hydraulic actuation (typical for excavator arms), contact dynamics with the ground during excavation and understanding of terrain deformation during the task. Finally, control requires precise ability to follow specified plans for safety as well as exact task execution. Developing all these components require detailed testing that is challenging in realistic environments. A method to alleviate this challenge is the use of realistic simulation. 


Many high-fidelity simulation environments focus on accurately modeling physics, such as simulating multi-body system  \cite{kurinov_automated_2020} and soil-tool interactions through FEE (Fundamental Earthmoving Equation) \cite{holz_real-time_2013}. 
Many classical controllers can be improved using data-driven methods, such as Reinforcement Learning (RL)-based controllers. These methods can incorporate not only motion of the vehicle, but can also be designed to solve high level goals. Generating sufficient data for RL models, however, requires conducting large-scale simulated experiments, which further emphasizes the need for scalable solutions. Similarly, modern autonomy often incorporates perception sensors like cameras, LiDARs, and RGB-D sensors. Achieving efficient control relies on the system’s ability to adapt to environmental changes, such as static and dynamic obstacles in real time. Yet, many excavator simulations focus on accurate kinematic and dynamic modeling and fail to provide the level of visual and structural fidelity necessary to fully leverage perception-based control strategies. Real-world excavators have limited sensing for various elements such as an inclinometer for the arm requiring sophisticated state estimation for accurate estimates of internal state over time. The errors in such state estimate have strong impact on excavation itself. These elements need to be modeled well in a simulator for the user to be able to develop realistic autonomous algorithms in simulation. Finally, designing an autonomy solution for excavation includes design questions such as number and placement of various sensors on an excavator, correct configuration and communication between the control system, sensors and compute, as well as realistic timing and interaction which all require a simulator to be able to simulate these aspects realistically for the simulator to be useful for the development of end-end autonomy solutions. 

Our work makes the following contributions
\begin{itemize}
    \item We have built \qq{} , the first comprehensive simulation environment to study autonomous excavation. \qq{} allows realistic modeling of perception, planning and control of excavators in realistic environments allowing the study of all aspects of excavation autonomy. 
    \item \qq{} is built to be customizable and extensible allowing users to import their excavators, environments and excavation tasks easily. With ROS integration, it is easy to quickly integrate existing autonomy modules as well. This allows users to quickly re-create their environment and focus on the development of autonomy algorithms. 
    \item Using Unity and AGX, \qq{} is scalable to use multiple excavators while performing high-fidelity simulation in real-time on desktop hardware. 
\end{itemize}

The simulation framework along with designed assets including robot and environment will be made available on the project website\footnotetext{\url{https://droneslab.github.io/tera}} on publication of this work. 

\section{Related Work}
Developing realistic solutions for autonomous robotics requires accurately modeling the physical properties, kinematics and rigid body dynamics of the robot. Additionally, for autonomous earthmoving, the mechanics of the soil and its interaction with said equipment also needs to be accurate. Prior works on autonomous excavation \cite{jud_heap_2021,osa_deep_2022,kurinov_automated_2020} utilize the model proposed by Park \cite{park_development_2002} to simulate the soil-tool interaction. This model starts from the fundamental earthmoving equation (FEE) \cite{reece_paper_1964} which enables fast real-time simulation due to its analytical nature but is limited to stationary conditions and does not describe the motion of the soil. The works \cite{holz_soil_2009,holz_real-time_2013,holz_advances_2015,jaiswal_deformable_2019} address this issue by combing FEE with particle dynamics to factor in motion by adaptively converting static soil into particles as the equipment deforms the terrain. However, as pointed out by \cite{jud_heap_2021}, these particle based simulations require heavy computations making them unsuitable for simulating multiple robots simultaneously for reinforcement learning. In contrast we use \cite{servin_multiscale_2021}, through an Algoryx (AGX) \cite{agx} plugin,  which is a computationally efficient model that uses a fast iterative solver to resolve the motion of the soil and a direct solver with high numerical precision for realistic forces and dynamics for the earthmoving equipment. This lets us simulate multiple excavators and not compromise on the accuracy of soil simulation

On the other hand, existing excavator simulators such as \cite{kwon_hydraulic_2008,gonzalez_3d_2009,ni_visual_2009,ni_design_2013,tao_low-cost_2008} and commercial solutions like the ones from Komatsu Ltd., \cite{noauthor_komatsu_nodate} are geared towards training human operators and not development of autonomous solutions. These simulators feature cockpits and virtual reality integrations to give human operators an extremely realistic experience but lack necessary features such as multi-excavator simulation, scripted interactions and data recording. \qq{} provides these through integrations with Robot Operating System (ROS)\cite{macenski_robot_2022}, the Unity Python API and completely open sourced code for custom modifications.


\section{The \qq{} Simulation Framework}
\label{sec:method}


\begin{figure}[h]
    \centering
    \includegraphics[width=1\linewidth]{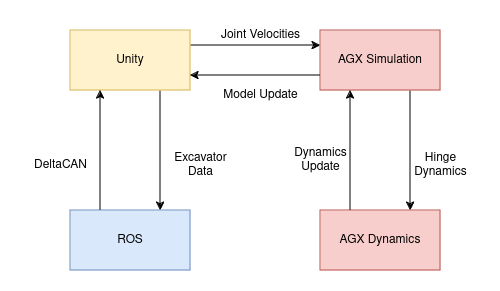}
    \caption{Communication Diagram of the Simulator}
    \label{fig:arch-diagram}
    \vspace{-10pt}
\end{figure}
\qq{} offers a highly-realistic simulation environment capable of modeling realistic excavators, environment and their interaction. \qq{} uses Unity3D for physics modeling and visualization. Unity3D is widely used in the robotics community due to it's powerful graphics rendering, physics
engines and development framework that aid in rapidly developing realistic, efficient and easily deployable
simulations. It also provides a physics engine to simulate real world physics such as rigid body kinematics, fluid dynamics and collisions. Interactions with the terrain occur using the AGX Dynamics Plugin \cite{agx_dynamics}. The AGX Terrain allows modeling and simulation of soil deformation via interactions between parts of the robot (tracks, bucket) and the soil. We can model physical properties of the terrain to adapt it to various real life terrains such as soil, gravel, sand and others. \autoref{fig:arch-diagram} shows the interaction between Unity, AGX and our simulator. 

From \autoref{fig:arch-diagram}, the Unity application commands joint velocities and updates the model through the AGX Dynamics. The dynamics engine computes the new state based on the defined link/joint properties. The advantage of using AGX is that it is able to perform realistic interaction simulation while being computationally efficient. For this, it only simulates the realistic physics in places where there is physical interaction and not the whole scene. This allows \qq{} to work in real-time while simulating multiple robots interacting with the world concurrently. 

\subsection{Creating Realistic Simulations in \qq{}}
\begin{figure}
    \centering
    \includegraphics[width=.9\linewidth]{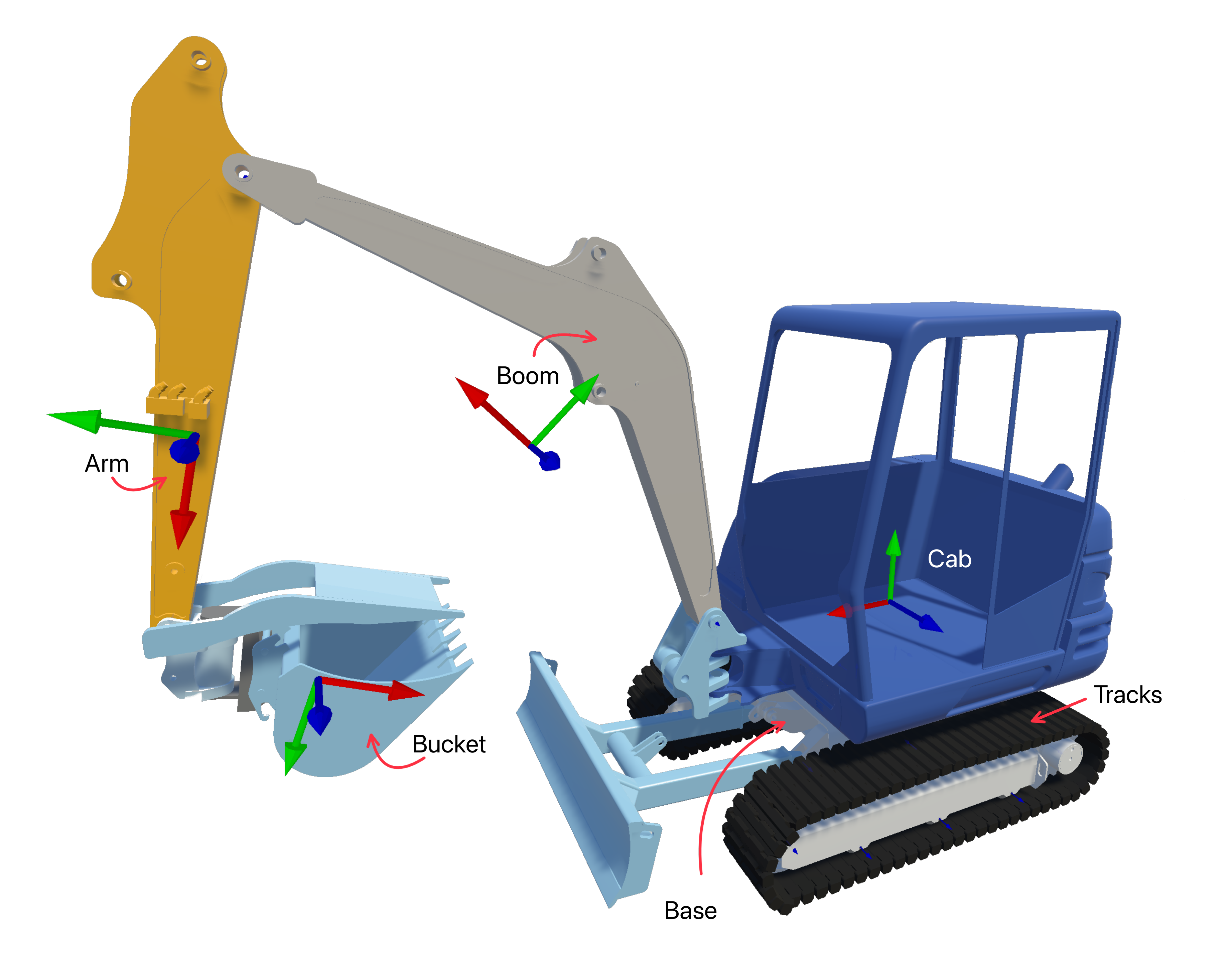}
    \caption{Rigid links origins for the excavator manipulator. X, Y, Z are Red, Green, Blue, respectively}
    \label{fig:excavator_iso}
    \vspace{-10pt}
\end{figure}

{\bf Robot Modeling: }
\label{sec:method-model}
\qq{} is capable of modeling excavators in great detail. The excavator model consists of 4 main components - a 4 degree-of-freedom manipulator, the cab, the base and the tracks. The manipulator is mounted on the cab which has the ability to slew freely about the base. The base is rigidly attached to the track mechanism which allows the excavator to move with a non-holonomic differential drive mechanism. \autoref{fig:excavator_iso} shows the modeled excavator with its links. For demonstration, we have modeled a Takeuchi TB-235 which is a compact, 4-ton excavator\cite{takeuchi}. Since the track interacts with the terrain it is defined as a custom AGX asset which includes properties such as thickness, material, number of links and the inter-link tension. But this is fairly representative of various excavators and a similar process can be followed to import other excavators.  We started by designing a CAD model,  established joints manually and set accurate dimensions. Once an accurate model of the excavator was designed in CAD, the STEP file was converted into a URDF using \cite{onshape_robot} which can be directly imported into Unity. 
{\bf Terrain Interaction: }
\label{sec:method-soil}
For simulation of a digging operation, the bucket and plow are defined as \texttt{shovel} elements in AGX. This element includes a top-edge, bottom-edge and cutting-edge. When the bucket makes contact with the terrain, the angle made by the cutting edge and the ground is evaluated. Based on the angle of the edge and the terrain shearing properties, the terrain deforms. When the terrain deforms, the excavated soil is converted to dynamic soil particles. These particles interact with other particles, terrain meshes and the excavator itself \cite{servin_multiscale_2021}. The mesh location from which the soil was excavated from now contains a depression indicating the removal of material. The underlying terrain data model consists of a 3D grid of cells, containing terrain data such as mass, compaction and soil type information. When a mass interacts with the terrain, the terrain properties and soil theory determine the type of changes \cite{desanto_compaction_1983}. The major properties include the Young's Modulus, Friction angle and Cohesion. Shear stresses and normal reactions for less cohesive materials causes early active terrain failures which reduces the effectiveness of the thrust generated by the track \cite{noauthor_terramechanics_2010}.


{\bf Sensor Simulation: }
\label{sec:method-sensors}
Sensors are integrated into \qq{}'s excavators via the Unity Sensors plugin \cite{unity_sensors}. The sensor suite includes RGB cameras, RGBD cameras, IMUs and LiDARs. Each sensor's internal parameters such as the camera's resolution, IMU acceleration bias/noise, etc. can be defined explicitly. In our sample excavator, we placed IMUs on the Chassis, Arm, Boom and Bucket as shown in \autoref{fig:excavator_iso}.  All configurations including exact pose can be defined in a YAML file that can be changed during the launch of the simulator. \autoref{fig:yaml} shows a sample configuraion with one chassis IMU.
    
\begin{figure}
    \footnotesize
    \begin{minted}[
    gobble=4,
    frame=single,
  ]{yaml}
    Excavator:
      - id: excavator1
        type: excavator
        offset: [1,1,1] 
        rotation: [0,0,0]
        sensors:
          - id: Chassis_IMU
            type: IMU
            topic: /imu_chassis
            location: CHASSIS
            noise: [0.1, 0.01]
            offset: [0.3436, 0.15, -0.2921]
            rotation: [0,-90,90]
    \end{minted}
    \caption{YAML configuration to spawn one IMU}
    \label{fig:yaml}
\end{figure}

{\bf Simulating Multiple Robots: }
\label{sec:method-multi}
Our configuration (YAML file) also allows the creation of multiple excavators, each configurable to specific needs. Each sensor within an excavator is assigned to that excavator's name-space, ensuring organization and data management. The configuration in ~\autoref{fig:yaml} will generate 1 excavator with the name \texttt{excavator1} having one IMU on the chassis. This configuration can then be extended to having multiple excavators with unique identifiers with a different sensor suite. 

\subsection{Implementation}
\label{sec:method-interface}



{\bf Realistic Actuation and Control: }
Once the URDF and the 3D Meshes are defined, it can be imported into Unity. Each link in the kinematic chain is defined with it's mass, inertia matrix (in X-Y-Z axis) and joint type. For revolute joints, a hinge constraint is defined between each link as they rotate about their Z axis. The constraint includes capabilities of modeling rotation limits, target speed controllers and friction blocks with defined compliance (stiffness), damping and force limits. A target speed controller is utilized to move the joints. Regardless of the input velocity command, the speed controller reaches the target speed with minimal transient response time. For the purpose of maintaining a model-agnostic nature of the proposed simulator, we allow the lower-level controller to  have nearly infinite acceleration, which is hardly achievable in real-world applications. To replicate the motion of a real excavator, we conducted an experimental study where a various range of input signals were commanded and joint velocity was monitored by the onboard inclinometers that provide angular position and velocity with respect to gravity. The actual velocity profiles are parameterized as:
\begin{equation}
    \omega^i(t) = \omega^i_{ss}(1+\sin(\eta^i t +\phi^{i} )e^{-\beta^i t})\label{eq:dynamics}
\end{equation}
 where $\omega^i_{ss}$ is the angular velocity of the joint at steady states, obtained from the experimental study and $i=$ [boom, arm, bucket]. The user-selected performance parameters, $\eta^i$, $\beta^i$ and $\phi^{i}$, govern the frequency of oscillation, decay rate and delay, respectively, which can be freely selected based on hardware performance.



{\bf ROS Bindings: }
The simulator provides relevant information about the state of the excavator using this framework as the backbone. Current position, orientation, and velocity of the cabin is provided using the \texttt{Odometry} msg. The angular position, angular velocity, and torque experienced by each joint in the arm is provided using the \texttt{JointState} msg and the relative 6DOF pose of the end-effector is provided with a \texttt{Transform} msg. Along with these, the mass in the bucket is also provided. \kar{describe frequency of these msgs and how control of the robot works using ROS.}



{\bf Emulating Control Hardware: }
Traditional excavators are pneumatic with a human controlling the various degrees of freedom through levers in the cabin. For example, a traditional TB235-2 excavator uses two 2-axis levers to control the slew, boom, arm, and bucket; two 1-axis levers to independently control the left and right tracks; one 1-axis lever to control the plow; and a pedal to control the arm’s swing \cite{takeuchi}. On our excavator, we have replaced the physical levers with controllable hardware that receives messages via the CAN bus. For simplicity, we send normalized values between  $[-1,+1]$ to control the various degrees of freedom. This interface is provided through a ROS wrapper via the \texttt{DeltaCAN} msg, a custom message encapsulating this control scheme. This data structure provides generalizability and can be replicated to other excavator configurations.

\section{Evaluation}

\subsection{Deformable Terrain / Soil Interaction}

\begin{figure}
    \centering
    \includegraphics[width=.9\linewidth]{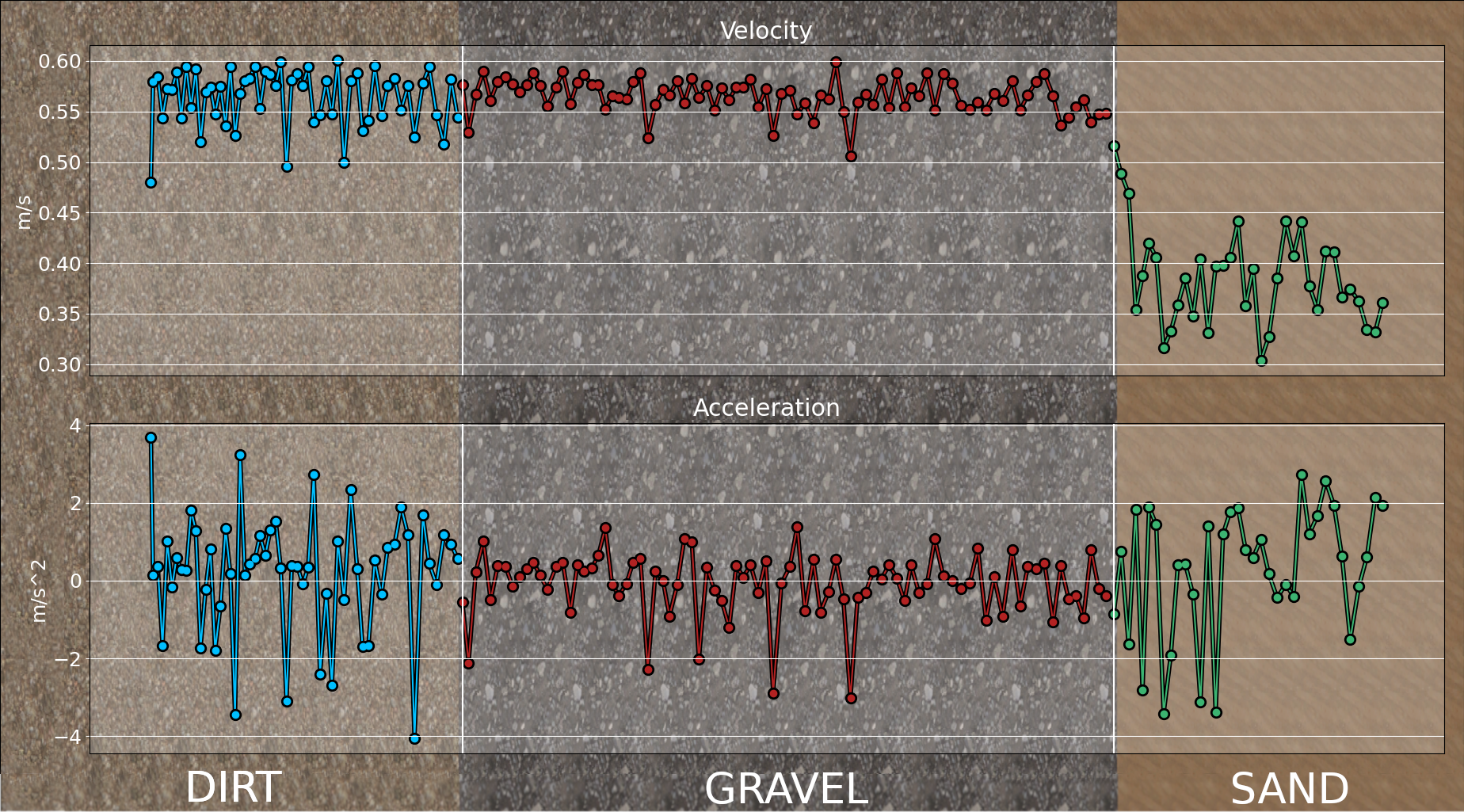}
    \caption{Velocity and Acceleration of Excavator traversing Dirt, Gravel and Sand in Simulation}
    \label{fig:terrains}
    \vspace{-10pt}
\end{figure}
Terrain such as dirt, gravel and sand exhibit different behaviors under stress. The spawned excavator deforms the environment based on it's own properties as well as the terrain's properties. \autoref{fig:terrains} shows the velocity and acceleration experienced by the simulated excavator in dirt, gravel and sand respectively. The Young's modulus are set to 6.5 MPa, 4.6 MPa and 4 MPa respectively. The bulk properties of dirt and gravel cause a significant reduction in the variance of both the velocity and acceleration of the excavator. A sharp decrease in speed is also observed when the excavator enters sand due to the reduction in the normal forces resulting in sinking.

\subsection{Bucket/Plow Interaction}



\begin{figure}
    \centering
    \includegraphics[width=\linewidth]{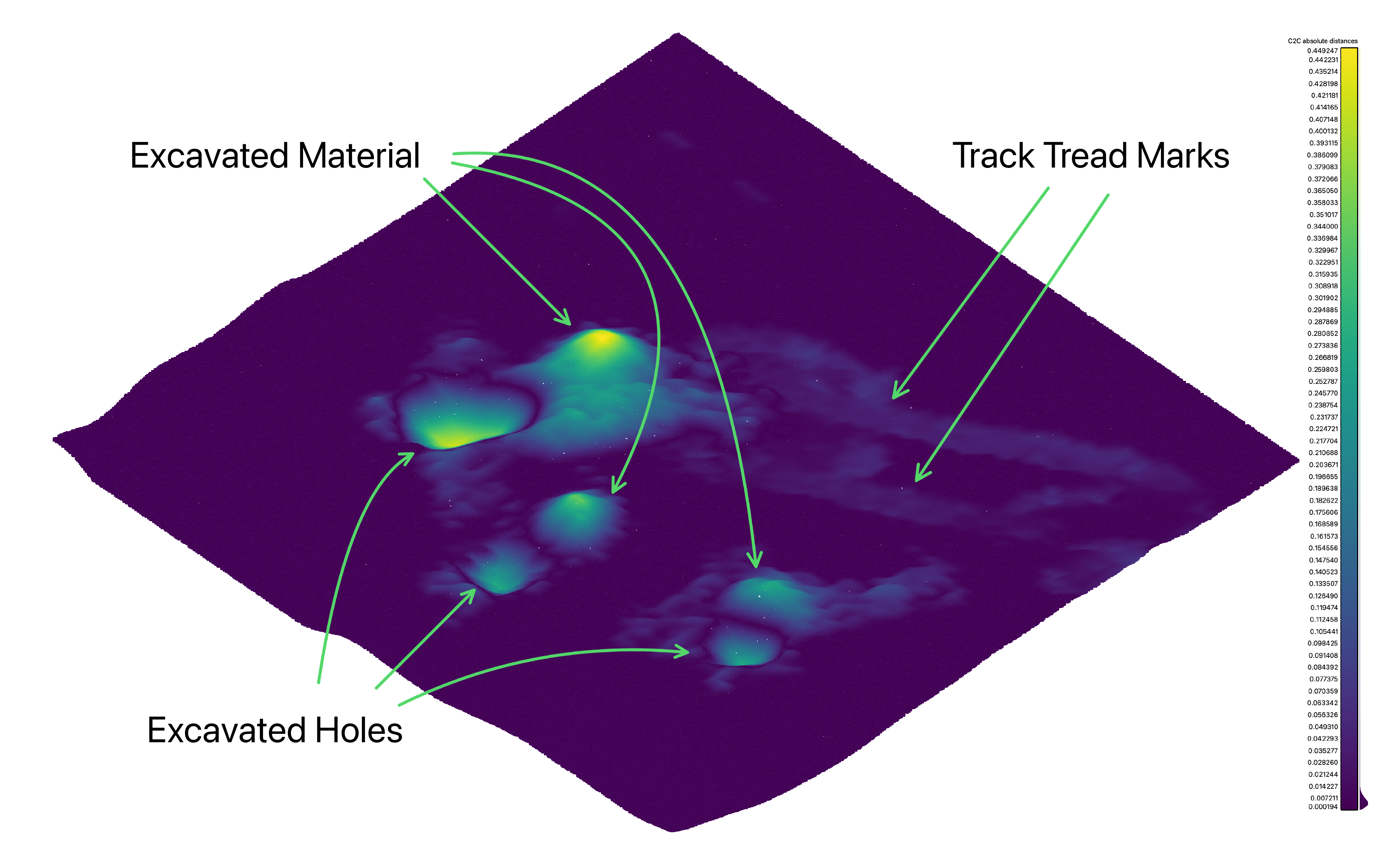}
    \caption{Isometric View of Terrain after 3 excavations }
    \label{fig:abs-terrain}
    \vspace{-10pt}
\end{figure}

\autoref{fig:abs-terrain} shows a digital elevation model view of the simulated terrain after performing digging motion. 
We can see that deformation is caused by both digging (the bucket interaction with terrain) as well as the motion (track interaction with terrain) of the excavator. Based on the terrain, such interaction simulation allows us to design realistic navigation and excavation algorithms that are cognizant of these properties. 

\subsection{Realistic Actuation and Manipulation}

\begin{figure}
    \centering
    \includegraphics[width=1\linewidth]{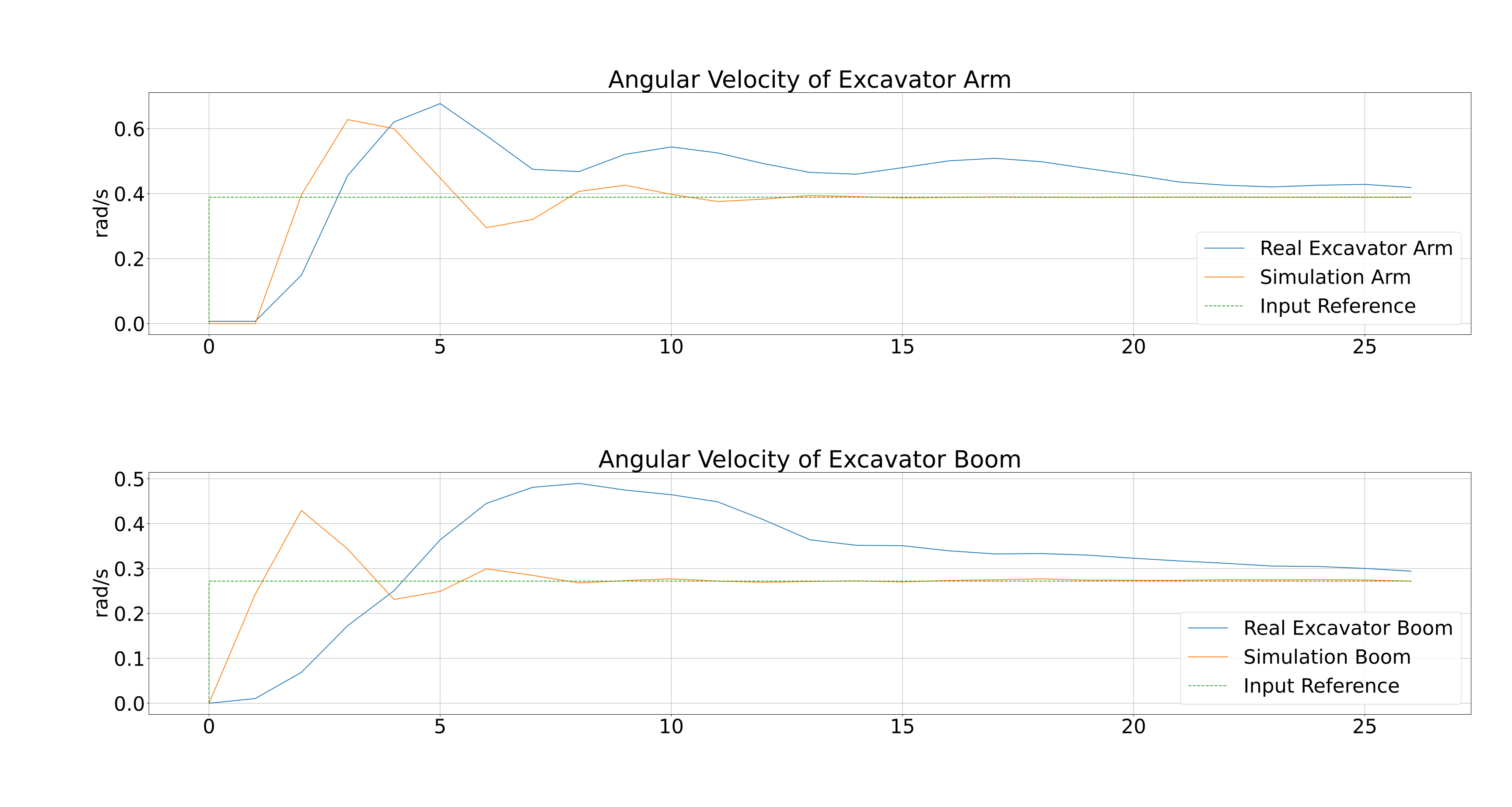}
    \caption{Angular Velocity Comparison of Excavator Arm/Boom in Real and Simulation with the reference signal}
    \label{fig:system_dynamics}
\end{figure}

\begin{figure}
    \centering
    \includegraphics[width=1\linewidth]{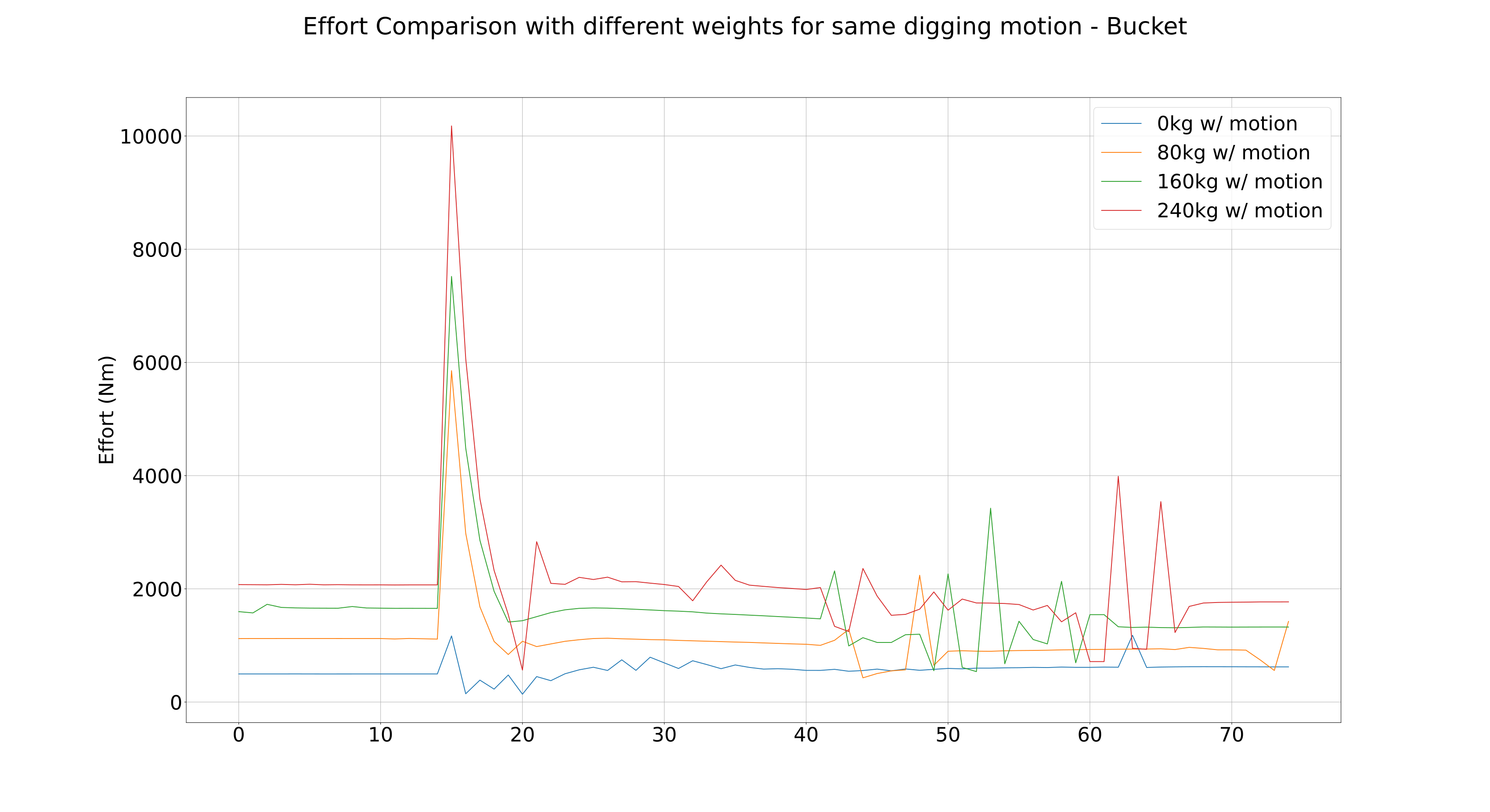}
    \caption{Effort on Bucket Joint (Simulation)}
    \label{fig:effort}
\end{figure}

Using \autoref{eq:dynamics}, we assign $\eta^i = 20$. $\beta^i = 6$ and $\phi^i = 0$ based on empirical testing.  \autoref{fig:system_dynamics} shows the angular velocity of the arm and boom recorded from the real excavator and the simulated excavator when an input of 60\% is provided. The simulation velocity tracks the real excavator velocity more accurately. The torque experienced by each joint is also provided in the \texttt{JointState} message. \autoref{fig:effort} shows the magnitude of the torque experienced by the bucket joint lifting varying masses. Real joints have a defined load/torque limit. Utilizing the manipulator outside these limits reduce the lifespan of the manipulator as well as incur additional utilization costs \cite{kot_method_2021}. Torque information can also be utilized to generate efficient trajectories for excavation tasks that are adaptable to differing soil conditions \cite{yang_time_2020}.
 
\subsection{Control}

\begin{figure}
    \centering
     \adjustbox{trim= 3.5cm 0cm 2cm 0.9cm, clip=True}{\includegraphics[width=1.8\linewidth]{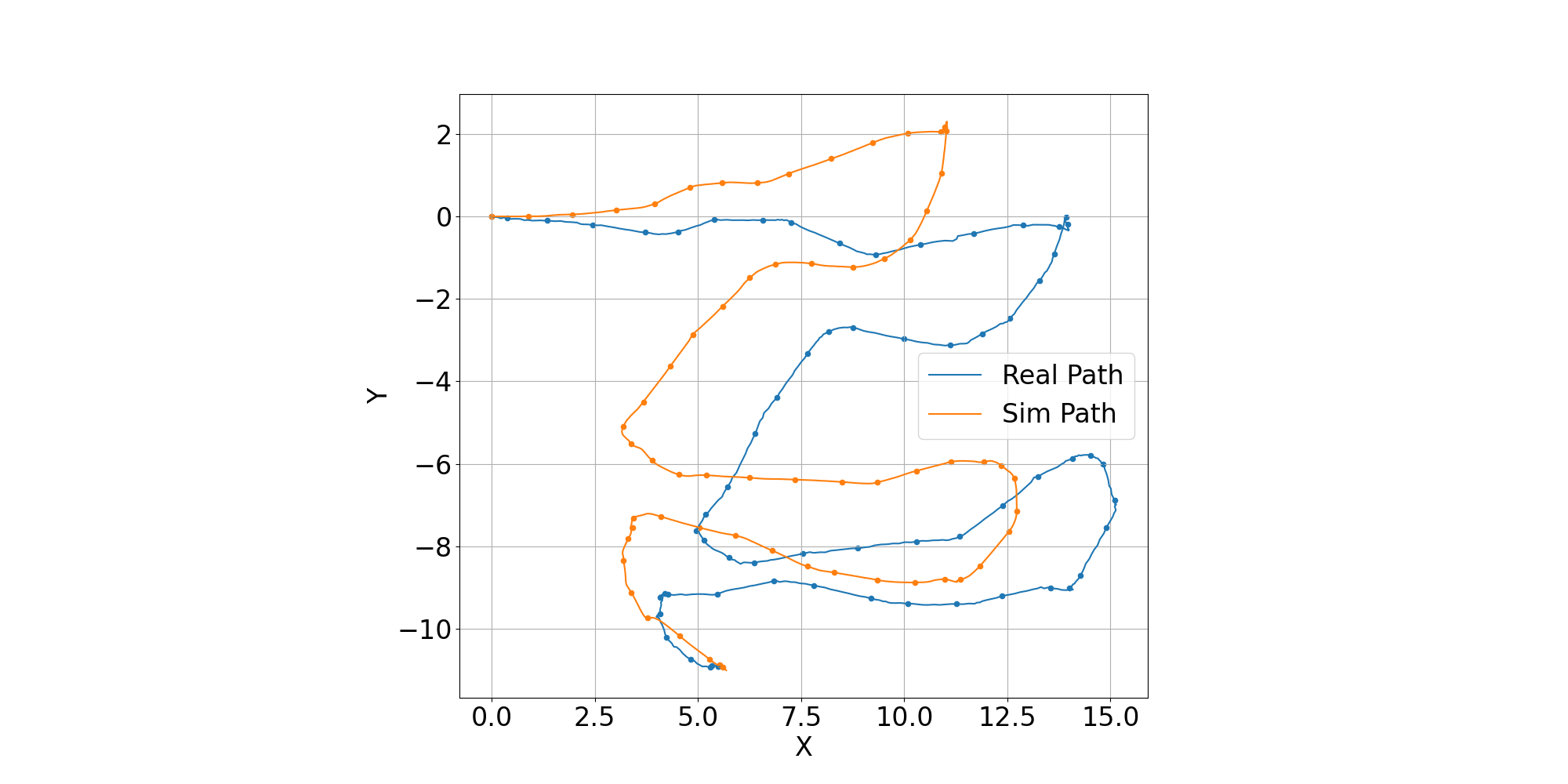}}
    \caption{Real/Sim Trajectory for the Same Controls}
    \label{fig:trajectory}
    \vspace{-10pt}
\end{figure}

To test the actuation of the tracks and the sim-to-real transfer capability of the simulator, the real excavator was controlled by a conventional joy input. These joystick inputs were recorded and replayed as input to the simulator (via ROS bag). The real excavator was equipped with 2 GPS receivers which provides accurate Latitude-Longitude-Altitude upto 0.75m. The Geodetic co-ordinates were converted to local ENU (Cartesian XYZ) co-ordinates using the Geodetic-ECEF-ENU conversion \cite{hofmann-wellenhof_global_2001} \cite{cai_unmanned_2011}. The path traced by the real excavator and the simulated excavator (obtained as ground truth from the Unity engine) is shown in \autoref{fig:trajectory}. Both trajectories are quite similar demonstrating realistic simulation of excavation navigation. The total track length of the real excavator is 57.39m while the length of the simulated excavator is 50.47m with an RMSE of 1.376m between them. Fine-tuning of the terrain parameters based on the real operational parameters could further reduce this gap.

\section{Discussion}

\subsection{Applications}
\qq{} has been specifically developed to address the autonomy challenges in excavation, earthmoving, and construction. In its current iteration, we believe it is a valuable tool to address  a broad range of problems in these areas.

\qq{} excels at enabling isolated tasks such as digging loose soil, picking and placing objects, and navigating uneven terrain. The simulation of soil interactions in \qq{} is more accurate and optimized compared to previous systems, which directly enhances the performance and reliability of these tasks. \qq{} goes beyond simple tasks by supporting more complex operations that require intricate coordination between the various actuated parts of an excavator. For example, the system can model behaviors like anchoring with the plow to gain leverage while digging, or using the thrust from the tracks to counteract forces on the bucket. Unlike previous models that focus either on the excavator’s manipulator for digging or its tracks for navigation, \qq{} offers a holistic approach by modeling the excavator as a full mobile manipulator. This integrated modeling, coupled with the full state information it provides, allows for the development of autonomy solutions capable of handling these advanced, multi-faceted interactions.

\begin{figure*}
    \centering
    \includegraphics[width=\linewidth]{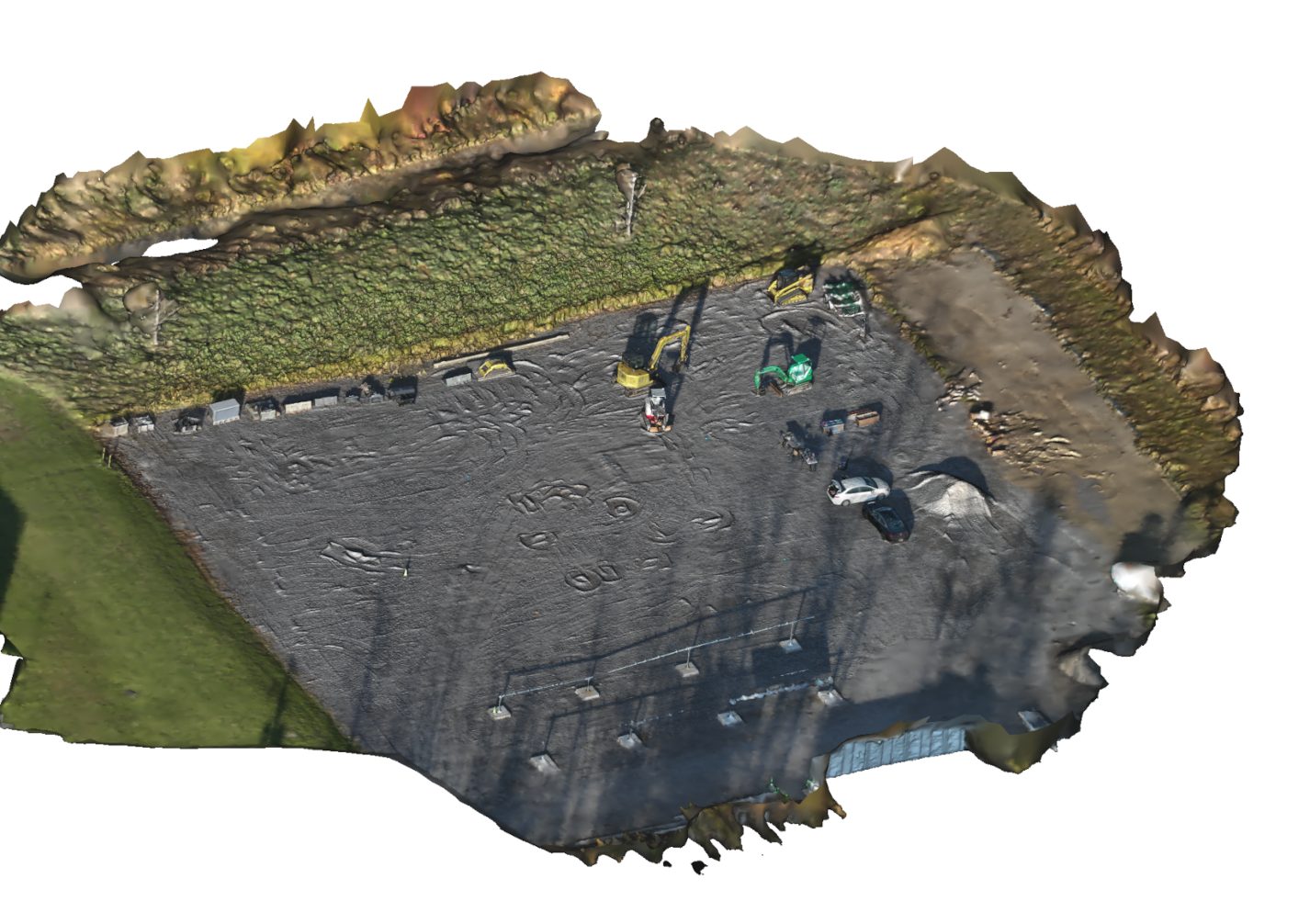}
    \caption{Recreation of our real-world environment in \qq{}}
    \label{fig:mesh-view}
\end{figure*}

\qq{} also allows tackling challenges related to perception and state estimation through its simulated sensors, including cameras, lidars and IMUs. 
Realistic simulation of these sensors allows for accurate development of decision making in dynamic and uncertain conditions. Finally, \qq{} supports the simulation of multiple excavators. This feature opens up opportunities for coordinating between multiple machines to jointly accomplish a task, a common use case in real-world operations. 
Together, these capabilities make \qq{} a powerful tool for advancing autonomous solutions in the excavation, earthmoving, and construction industries.

\subsection{Limitations}

The lower level dynamics engine is not exposed to the end user. The end user can specify a target speed to a mid-level controller which is then converted to the lower level control. The dynamics of the excavator are approximated using kinematics. Additional parameters such as hydraulic fluid properties, inter-link friction, etc. are not accounted for. The AGX Dynamics simulation block that handles the interactions uses 2 dedicated CPU threads to perform computations and is not scaled to the number of excavators. The processing time of the simulator is also dependent on the size of the terrain as well as the total number of dynamic particles during any interaction. Excavating large amounts of the terrain results in a significant decrease in the performance. Usage of the simulated camera is limited due to the high computation cost of rendering environments with photo-realistic elements. We expect to address several of the control and dynamics challenges in future editions of \qq{}. 

\section{Conclusion}

In this work we present \qq{}, a simulation framework built on Unity/AGX for realistic end-end simulation of excavators. Through detailed modeling, we describe ways the user can recreate their own excavator and environment in \qq{}. Key features of \qq{} include deformable terrain, realistic perception, customizable control, and easy configuration for customizability. Through micro-benchmarks and end-end simulations, we demonstrate \qq{} simulating a representative excavator and its environment along with realistic interaction. The simulation framework and assets will be made available on the project page on publication. 




\bibliographystyle{ieeetr}
\bibliography{references_URL}
\end{document}